\documentclass{article}

  \usepackage[preprint, nonatbib]{neurips_2021}

\usepackage[utf8]{inputenc} % allow utf-8 input
\usepackage[T1]{fontenc}    % use 8-bit T1 fonts
\usepackage{hyperref}       % hyperlinks
\usepackage{url}            % simple URL typesetting
\usepackage{booktabs}       % professional-quality tables
\usepackage{amsfonts}       % blackboard math symbols
\usepackage{nicefrac}       % compact symbols for 1/2, etc.
\usepackage{microtype}      % microtypography
\usepackage{xcolor}         % colors

\usepackage{algorithm, algorithmic}
\usepackage[pdftex]{graphicx}
\usepackage{amsmath}
\usepackage{xspace}
\def\sysname{\text{SIEGE}\xspace}
\usepackage{comment}

\title{Self-supervised Incremental Deep Graph Learning for Ethereum Phishing Scam Detection}

\author{%
  Shucheng Li, Fengyuan Xu\thanks{Corresponding author.}, Runchuan Wang, Sheng Zhong\\
  National Key Lab for Novel Software Technology, Nanjing University, China \\
  \texttt{shuchengli@smail.nju.edu.cn}
}
\begin{document}

\maketitle

\begin{abstract}
In recent years, phishing scams have become the crime type with the largest money involved on Ethereum, the second-largest blockchain platform. Meanwhile, graph neural network (GNN) has shown promising performance in various node classification tasks. However, for Ethereum transaction data, which could be naturally abstracted to a real-world complex graph, the scarcity of labels and the huge volume of transaction data make it difficult to take advantage of GNN methods. Here in this paper, to address the two challenges, we propose a \textbf{S}elf-supervised \textbf{I}ncr\textbf{E}mental deep \textbf{G}raph l\textbf{E}arning model (\textbf{SIEGE}), for the phishing scam detection problem on Ethereum. In our model, two pretext tasks designed from spatial and temporal perspectives help us effectively learn useful node embedding from the huge amount of unlabelled transaction data. And the incremental paradigm allows us to efficiently handle large-scale transaction data and help the model maintain good performance when the data distribution is drastically changing. We collect transaction records about half a year from Ethereum and our extensive experiments show that our model consistently outperforms strong baselines in both transductive and inductive settings.
%Furthermore, we provide an anchor nodes-based sampling method to accelerate model convergence.
\end{abstract}

%%self-supervised incremental learning for anomaly detection on large-scale transaction graphs

\section{Introduction}
\label{sec:intro}

Ethereum~\cite{wood2014ethereum}, one of the most popular and scalable blockchains, has reached \$28 billion market cap so
far. It unleashes the full potential of smart contracts which automatically manage and approve transactions without a
centralized entity, and thus attracts a lot of attention and resources via its various decentralized finance (DeFi)
applications atop of smart contracts. However, the thriving of Ethereum also puts millions of its users at risk of
malicious attacks~\cite{huang2021survey}. Especially, many of these attacks, such as the phishing
scam~\cite{chen2020phishing}\cite{Chen2020PhishingSD}\cite{wang2021tsgn}\cite{wu2020phishers}\cite{yuan2020detecting}
and Ponzi scheme~\cite{bartoletti2020dissecting}\cite{chen2018detecting}, aim to extort money from their victims in
Ethereum, which is hard to defend with the software security approaches or smart contract analysis alone. Therefore, intelligent analytics at the transactional activity level is necessary.

Recently, deep graph
learning\cite{perozzi2014deepwalk}\cite{tang2015line}\cite{grover2016node2vec}\cite{kipf2016semi}\cite{hamilton2017inductive}\cite{velivckovic2018deep}
has shown its superior ability to learn good node representations for graphs, which is useful to downstream tasks.
Since all transactional activities in Ethereum can naturally form a large-scale transaction graph (formulation details
are described in Section \ref{sec:related_eth}), the effectiveness of security analytics could be greatly improved
if leveraging the advantages of deep graph learning methods on Ethereum. Unfortunately, there are two key challenges
when introducing deep graph learning on such a large-scale transaction graph for the task category of anomaly
detections.

The first challenge is the \textbf{data scalability}. The size of the Ethereum transaction graph is extremely large, with over 1,151,443,821 transactions (excluding internal transactions) so far. This graph continues evolving by adding new transactions every second. After the deployment of Ethereum 2.0 in 2022, there will be thousands of transactions generated in a second. The other challenge is the \textbf{label scarcity}. It is in general hard to
obtain positive samples for the training of anomaly detections. Moreover, it is usually hard to find labeled data in
recently-generated graph parts which need inspections most, creating an issue of temporal labeling imbalance.

Existing works~\cite{chen2020phishing}\cite{wu2020phishers}\cite{yuan2020detecting} are not able to handle large
evolving graph and few positive labels. Instead, most of them alleviate these two challenges via the graph sampling mechanism.
The whole graph is shrunk to a proper computable size by picking nodes and edges related to labeled positive data.
However, such a biased sampling process could lead to serious data distribution distortion and still suffers from the
issue of temporal labeling imbalance. Additionally, most of these methods are transductive and not suitable to inductive
scenarios like Ethereum and all other evolving graphs.

In this paper, we attempt to address in a different way the challenging anomaly detection tasks on large-scale evolving
graphs. \textbf{First}, we would like to take advantage of all data we have, rather than getting rid of them. The self-supervised
learning (SSL) path is taken in our work because it can dig out more information through the supervision signals
directly from data themselves without worrying about annotations. The SSL effectiveness has been proven by many successful cases
in natural language processing, computer vision, and graph data~\cite{jing2020self}. However, existing unsupervised methods on
graphs~\cite{perozzi2014deepwalk}\cite{tang2015line}\cite{grover2016node2vec}\cite{kipf2016variational}\cite{hamilton2017inductive}\cite{velivckovic2018deep}
cannot directly apply onto the large-size graph, and their pretext tasks cannot fully capture both spatial and temporal
dynamics in the graph evolving. \textbf{Second}, we therefore would like to design an incremental learning paradigm for such a large and evolving graph category which represents many real-world graph learning scenarios. \textbf{Last}, we would like to make
our method inductive, and use the anomaly detection task of Ethereum phishing scam as an example to demonstrate the
effectiveness of our work to the downstream tasks in blockchains.

Therefore, we propose a \textbf{S}elf-supervised \textbf{I}ncr\textbf{E}mental deep \textbf{G}raph l\textbf{E}arning
model (\textbf{SIEGE}) for anomaly detection tasks like phishing scam detection on large scale evolving graphs like
Ethereum transaction graph. Instead of learning the whole graph at once, \sysname splits the transaction graph into
pieces of suitable size (i.e. a set of Ehereum transaction blocks) and incrementally consumes them one by one in the
temporal order to learn the node representation. The intermediate learning results of one graph piece are passed by our
design to the next piece to enable incremental computing. In this way, the size of the in-memory graph is containable
and adaptive to changes up to date. For the learning inside a graph piece, to address the label scarcity, we employ self-supervised learning to
generate the node embedding with an unsupervised representation learning approach, and design two spatial and temporal
pretext tasks to ensure that the final node representations are as informative as possible, with respect to the phishing
scam detection task.

To demonstrate the effectiveness of \sysname, we collected transaction data of about half a year on Ethereum to build the graph dataset (more than 70 million transactions), and then extensive experiments conducted in both transductive and inductive settings show that \sysname consistently outperforms baselines by 4\% \textasciitilde 16\%  in F-1 score.

\section{Related work}
We mainly introduce two research directions related to this work here: 1) Ethereum phishing scam detection; 2) deep graph learning. For 1), we will show some background knowledge about the basics of Ethereum and the Ethereum transaction graph, and then describe the methodology of existing works for phishing scam detection on Ethereum. We will also make some comparisons about the difference between Ethereum and other blockchain platforms (especially Bitcoin). For 2), we will first introduce some basic GNN methods. Then, we will present the self-supervised learning for graphs.

\subsection{Ethereum phishing scam detection}
\label{sec:related_eth}

\paragraph{Ethereum basics.} Inspired by Bitcoin\cite{nakamoto2019bitcoin}, Ethereum\cite{wood2014ethereum} aims to become a blockchain platform of the next generation that supports both the cryptocurrency and decentralized applications (Dapps). The key is the designed smart contracts on Ethereum, which can be easily deployed in a transaction and execute preset functions when called by another account. It provides convenience for users because it allows for trusted transactions between users without a third party. Specifically in Ethereum, two types of accounts existed: external accounts controlled by users and contract accounts with code stored together. If we abstract the account as a node and the transaction as an edge, then we can get a transaction graph. It should be added that for Bitcoin, there is no concept of "accounts" or "balance", the basic block of Bitcoin is the unspent transaction output (UTXO), and the transaction in Bitcoin can have multiple outputs and inputs, while the transaction in Ethereum is one-on-one. This is part of the reason why we studied the Ethereum transaction graph. \cite{chen2020understanding} firstly employs graph analysis to study the characteristics (degree distribution, clustering coefficients, etc.) of transaction graphs on Ethereum and proposes a rule-based anomaly detection method. It provides some insights into the behavior pattern of accounts in Ethereum.

\paragraph{Phishing scams in Ethereum.} The thriving Ethereum market has also bred a lot of criminal activities like phishing scam\cite{huang2021survey}\cite{chen2020phishing}\cite{Chen2020PhishingSD}\cite{wu2020phishers}, Ponzi scheme\cite{bartoletti2020dissecting}\cite{chen2018detecting}, ransomware\cite{delgado2020blockchain}, etc. For the phishing scam detection problem on Ethereum, there are two main categories of existing methods, the former mainly employs shallow models such as 1) traditional machine learning methods with dedicated feature engineering\cite{Chen2020PhishingSD}, and 2) some random-walk based network embedding methods\cite{wu2020phishers}, such as DeepWalk\cite{perozzi2014deepwalk}, Node2Vec\cite{grover2016node2vec}, etc. The latter applies some deep graph models like the graph convolutional network\cite{kipf2016semi} (GCN) based methods. These methods show pretty good performance in their data collected. However, as mentioned in Section \ref{sec:intro}, two problems limit the practical application of these methods. 1) Most of the existing methods are transductive, which make predictions in a single fixed graph. Therefore, they do not naturally generalize to unseen nodes or sub-graphs, and for a high-throughput active blockchain platform like Ethereum, an inductive method to process fast-evolving in-coming transaction data is in need. 2)  A biased sampling process is employed in graph data generation. To alleviate the label scarcity problem and scalability problem, existing methods\cite{chen2020phishing} generate the graph data by a random walk sampling process with labeled phishing nodes as the start, which will result in different node distribution between the sampled graph and original graph. Hence, in this paper, we propose \sysname to learn useful node representations from the original transaction without sampling. It is also inductive and can be used for phishing node detection in an entirely new transaction graph.

\subsection{Deep graph learning}
\paragraph{Graph neural networks.} Deep learning has been a great success in computer vision, speech recognition, and natural language processing due to the powerful representation learning capability. Recently, increasing attention is paid to how to apply deep learning for non-euclidean data like graphs, which is widely used in various tasks, such as social network\cite{fan2019graph}, protein interface prediction\cite{fout2017protein}, knowledge graph embedding\cite{wang2019knowledge}, etc. In retrospect, graph neural networks also help with lots of tasks in computer vision or natural language processing, like object detection\cite{shi2020point}, action recognition\cite{shi2019skeleton}, machine translation\cite{beck2018graph}\cite{bastings2017graph}, semantic parsing\cite{bogin2019representing}\cite{li2020graph}\cite{xu2018exploiting}, etc. Specifically, most GNNs exploit a neighborhood aggregation mechanism that could incorporate information from both the node attributes and graph topology. The node embedding generated can then be used for downstream tasks such as node classification, link prediction, graph classification (an extra readout module in need), etc.

\paragraph{Self-supervised learning for graphs.} Self-supervised learning (SSL) has been proved pretty useful when a large volume of unlabelled data is available\cite{devlin2018bert}\cite{chen2020simple}. Compared to supervised learning usually with manual annotations used as groundtruth, supervised learning aims to acquire the groundtruth from the data itself by different pretext tasks, which could somehow alleviate the poor generalization resulted from over-fitting, and weak robustness faced with adversarial attacks\cite{liu2020self}. For SSL in graph, the inherent topology indicates that the nodes are not independent, which makes it difficult for direct use of existing frameworks\cite{jing2020self}. However, rich structural information in graphs could also help the design of the pretext tasks. In a nutshell, the design of the pretext task remains the crucial point for SSL in graphs. Existing work of SSL\cite{you2020does}\cite{rong2020self}\cite{jing2020self} in graph mostly focus on simple graphs, i.e., homogeneous attributed graphs in a relatively small size. And how to apply SSL on large-scale complex graphs remains to be studied. To the best of our knowledge, we are the first work that proposes a new pretext task for a large-scale temporal graph.

\section{Methodology of SIEGE}

In this section, we will introduce each module of \sysname. In Section \ref{sec:ssl}, we will present how to design specific pretext tasks, which are divided into two parts, spatial pretext task, and temporal pretext task. And in Section \ref{sec:inc}, we will discuss the usage of the incremental learning module in \sysname and then give an overview of \sysname.

\subsection{Self-supervised training}
\label{sec:ssl}
Before performing self-supervised learning, we assume that the original graph data $\mathcal{G}$ has been divided into $N$ splits: $\{ \mathcal{G}_1, \mathcal{G}_2, \cdots, \mathcal{G}_N \}$ in the order of block number. For graph split $i$, $\mathcal{G}_i = (\mathcal{V}_i, \mathcal{E}_i, \mathbf{X}_i)$, where $\mathcal{V}_i = \{ v_1, v_2, \cdots, v_{N_i} \}$ is the set of $\mathcal{N}_i$ nodes, $\mathcal{E}_i$ stands for the edge set and $\mathbf{X}_i$ is the node feature matrix. Moreover, we use $\mathbf{A}_i$ as the adjacency matrix of graph split $i$, where $\mathbf{A}_{i,m,n} = 1$ if there is a edge between node $v_{i,m}$ and $v_{i,n}$ else $\mathbf{A}_{i,m,n} = 0$. To conclude, each graph split is a directed graph with node attributes and without labels. The objective of our SSL module can be abstracted to minimize the the pretext task loss $\mathcal{L}_{pretext}$,
\begin{equation}
    \min_{g} \mathcal{L}_{pretext}(\mathcal{A}, \mathbf{X}, g) = \sum_{v_i \in \mathcal{V}_{pretext}} \mathcal{D}(g(\mathcal{G})_{v_i}, y_{pretext_i})
\end{equation}
where $g$ is the GNN encoder (feature extractor) and the $y_{pretext_i}$ stands for the groundtruth of node $v_i$ acquired by pretext task. $\mathcal{G}$ is one of the graph splits. In addition, $\mathcal{V}_{pretext}$ is the node set used in the pretext task and the discriminator $\mathcal{D}$ is used to measure the relationship between the node embedding of $v_i$ and $y_{pretext_i}$. It should be noted that our self-supervised model involves only node-level pretext task, for sub-graph or graph-level pretext task, we leave them for future work. After pre-training with the pretext task, the parameters of GNN encoder $g$ will be frozen to be used for generation of node representation $\mathbf{Z} = \{ \mathbf{z}_1, \mathbf{z}_2, \cdots, \mathbf{z}_{N_i} \}$, which will be fed into the final classifier for phishing node prediction downstream task.

Since the original graph has been split into multiple pieces by block number (equivalently, by time), we firstly design a spatial pretext task for each graph piece to learn the distance information between nodes. Then a temporal pretext task is applied to learn the relationship between different graphs.

Intuitively, for the phishing scam detection problem, a well-designed pretext task should meet the following prerequisites: 1) The extracted data labels should reflect the characteristics of the data itself. 2) Due to the large scale of Ethereum transaction data, it should be possible for users to obtain the extracted data labels with relatively low time complexity, otherwise, the time cost for pre-training would be unacceptable. 3) The pretext task could incorporate some domain knowledge, but it should not be overly detailed. Or else, the applicability of the task would be limited and vulnerable to adversarial attacks.

\paragraph{Spatial pretext task.}

\begin{figure}[ht!]
    \centering
    \includegraphics[width=1.0\linewidth]{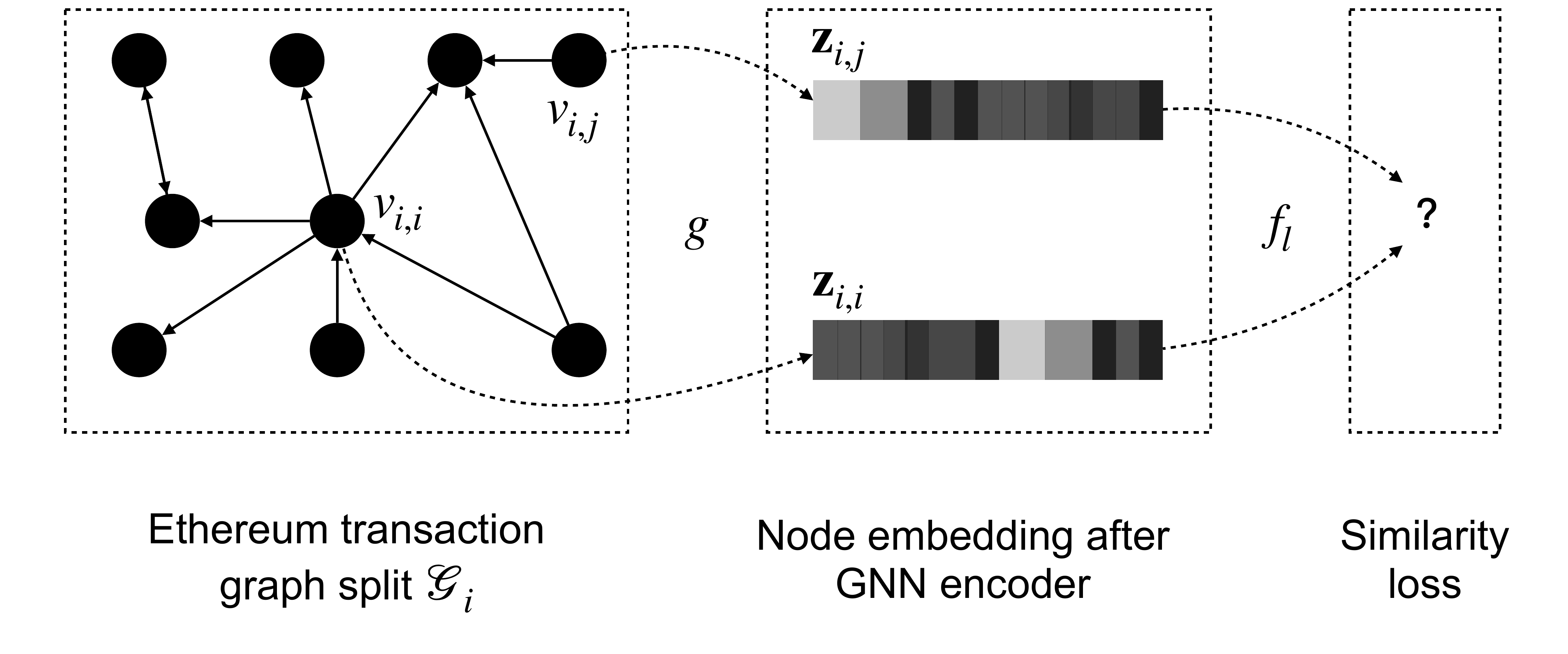}
    \caption{An overview of the spatial pretext task in \sysname. Where $g$ is the GNN encoder and $f_l$ is a linear transformation layer.}
    \label{fig:spatial_task}
\end{figure}

For the nodes in the Ethereum transaction graph, we believe that the nodes with frequent transactions should have a similar node representation. To efficiently capture this kind of spatial relationships (neighborhood relationships) between nodes in the graph, we design a spatial pretext task, as shown in Figure \ref{fig:spatial_task}. First, for a certain graph split $\mathcal{G}_i$, we firstly calculate its node embedding $\mathbf{Z}_i = \{ \mathbf{z}_{i,1}, \mathbf{z}_{i,2}, \cdots, \mathbf{z}_{i,N_i} \}$ by GNN feature extractor. Then, for node $v_{i,i}$, if node $v_{i,j}$ is a k-hop neighbor of $v_{i,i}$ (here we do not consider the edge direction), we maximize the similarity between them. Vice-versa, we let their node representations as dissimilar as possible. Formally,
\begin{equation}
\label{eq:spatial}
\begin{aligned}
    \mathcal{L}_{spatial}(\mathcal{A}, \mathbf{X}, g) = &\frac{1}{| \mathcal{P}_{spatial_i} |}\sum_{(v_{i,i}, v_{i,j}) \in \mathcal{P}_{spatial_i}} -f_{sim}(f_l(\mathbf{z}_{i,i}), f_l(\mathbf{z}_{i,j}))
    \\
    + &\frac{1}{| \mathcal{\overline{P}}_{spatial_i} |}\sum_{(v_{i,i}, v_{i,k}) \in \mathcal{\overline{P}}_{spatial_i}} f_{sim}(f_l(\mathbf{z}_{i,i}), f_l(\mathbf{z}_{i,k}))
\end{aligned}
\end{equation}

Where $\mathcal{P}_{spatial_i}$ and $\mathcal{\overline{P}}_{spatial_i}$ denote the set of node pairs in and not in the k-hop neighborhood for $\mathcal{V}_i$, respectively. $f_{sim}$ is the similarity function and we exploit $f_{sim}(a,b) = \sigma(a^Tb)$ here. $f_l$ is a linear transformation layer and $\mathbf{z}_{i,j}$ is the node embedding of $v_{i,j}$, which is the node $j$ in $\mathcal{G}_i$.

\paragraph{Temporal pretext task.}

\begin{figure}[ht!]
    \centering
    \includegraphics[width=1.0\linewidth]{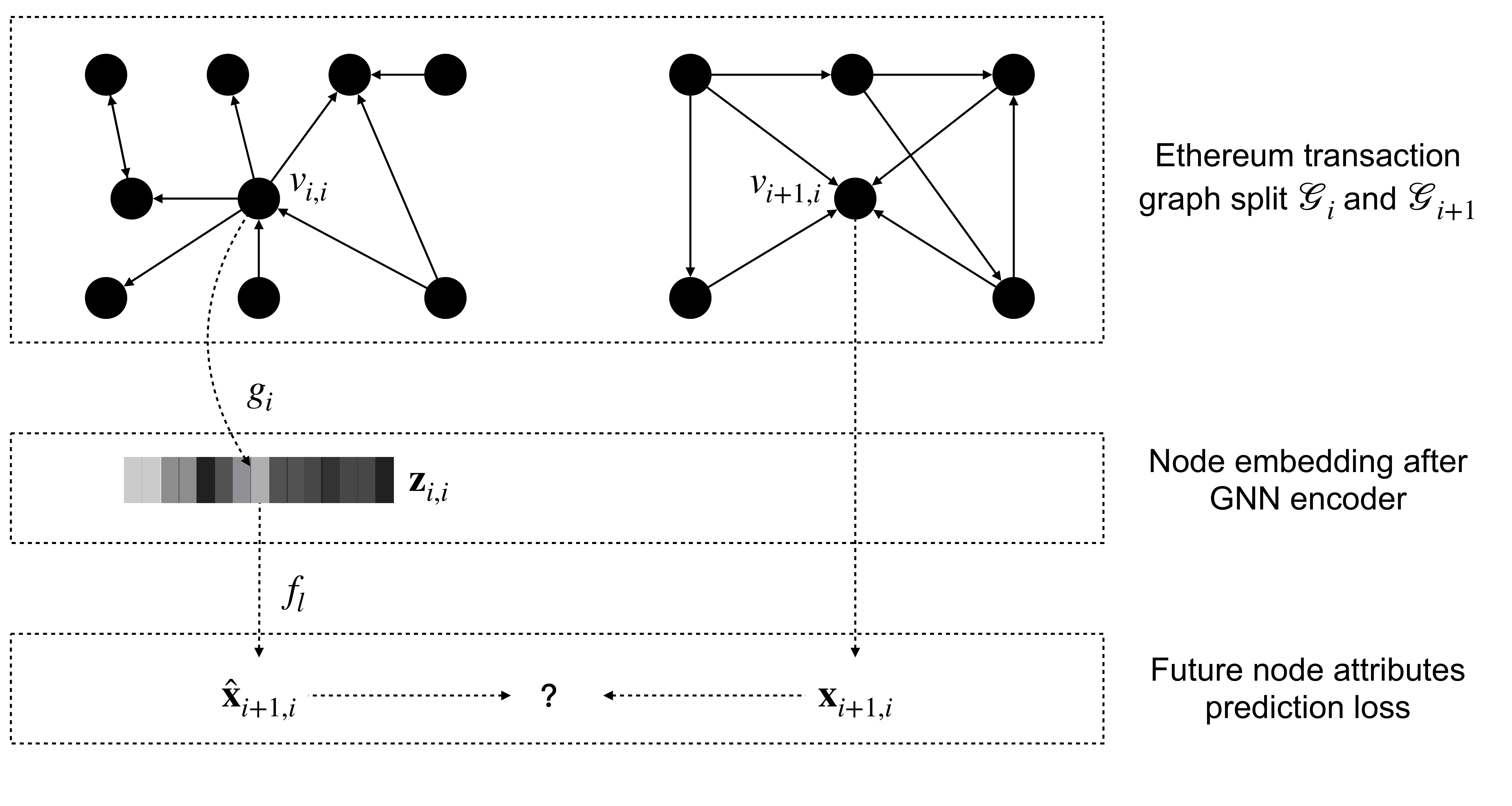}
    \caption{An overview of the temporal pretext task in \sysname. Where $g_i$ is the GNN encoder for the i-th graph split and $f_l$ is a linear transformation layer.}
    \label{fig:temporal_task}
\end{figure}

Since the original graph has been divided into $N$ splits and we learn the spatial information in each of them with the pretext task above. However, it is also important for us to consider the relationship between different graph splits while modeling each of them. Therefore, we design a temporal pretext task as follows.

First, since each graph split contains a large amount of transaction data, and the behavior of each account has continuity in the time dimension, a certain percentage of nodes (accounts) must overlap between two adjacent splits. This set of nodes forms a bridge between two adjacent graph splits, and we use this overlap node-set $\mathcal{V}_o$ to model the relationship between two graph splits in our temporal pretext task. The overview of the temporal pretext task has been shown in Figure \ref{fig:temporal_task}. Formally, we minimize the loss $\mathcal{L}_{temporal}$,

\begin{equation}
\label{eq:temporal}
    \mathcal{L}_{temporal}(\mathcal{A}, \mathbf{X}, g) = \frac{1}{| \mathcal{V}_{o_i} |}\sum_{v_{i,i} \in \mathcal{V}_{o_i}} \| f_l(\mathbf{z}_{i,i}) - \mathbf{x}_{i+1, i} \|^2
\end{equation}

Here, $\mathcal{V}_{o_i}$ is the overlap node set between graph split $\mathcal{G}_i$ and $\mathcal{G}_{i+1}$. And $\mathbf{x}_{i+1, i}$ is the node attributes of node $v_{i+1, i}$, which points to the same account as node $v_{i,i}$.

It is important that for a node, the transactions that occur in split $i$ will influence the behavior of this node in the next split. And our experiments show that this temporal pretext task can indeed help us learn this connection.

\subsection{Incremental training}
\label{sec:inc}
Due to the huge and rapidly increasing size of Ethereum trading data, it is not practical to feed all transaction data into the model at once. It is why we utilize an incremental training method by feeding partial training data at a time, which alleviates the data scalability problem on the one hand, and helps the model adapt to new trading data distributions on the other hand, as Ethereum market changes are more dramatic than traditional trading platforms.

In the following, we will briefly conclude \sysname to several steps with both the self-supervised learning module in Section \ref{sec:ssl} and incremental training in this section.

\begin{enumerate}
    \item For $\mathcal{G} = \{ \mathcal{G}_1, \mathcal{G}_2, \cdots, \mathcal{G}_N \}$, with each $\mathcal{G}_i = (\mathbf{A}_i, \mathbf{X}_i)$.
    \item In a graph split $\mathcal{G}_i$. Let GNN encoder $g_i = g_{i-1}^{\prime\prime}$, which is the updated GNN encoder in $\mathcal{G}_{i-1}$ after pre-training. For $\mathcal{V}_{o_{i-1}}$, the set of nodes which have appeared in both $\mathcal{G}_{i-1}$ and $\mathcal{G}_{i}$, obtain their final node embeddings $\mathbf{Z}_{o_{i-1}}^{\prime\prime}$ after pre-training in $\mathcal{G}_{i-1}$.
    \item Obtain node attributes $\mathbf{X}_i$. Let $\mathbf{Z}_{o_{i-1}}^{\prime\prime}$ go through a linear transformation and concatenate them with $\mathbf{X}_i$, obtain $\mathbf{X}_{concat_i}$. For node $v \notin \mathcal{V}_o $ or when $i==1$, concatenated with zero vector.
    \item Obtain node embeddings $\mathbf{Z}_i = g_i(\mathbf{A}_i, \mathbf{X}_{concat_i})$ through a GNN encoder $g_i$.
    \item With node embeddings $\mathbf{Z}_i$, GNN encoder $g_i$, minimize the spatial pretext task loss $\mathcal{L}_{spatial}$ in Equation \ref{eq:spatial}. Update parameters of $g_i$.
    \label{step:5}
    \item Obtain new node embeddings $\mathbf{Z}_i^{\prime} = g_i^{\prime}(\mathbf{A}_i, \mathbf{X}_{concat_i})$ through a updated GNN encoder $g_i^{\prime}$ in Step \ref{step:5}.
    \item With new node embeddings $\mathbf{Z}_i^{\prime}$, updated GNN encoder $g_i^{\prime}$, minimize the temporal pretext task loss $\mathcal{L}_{temporal}$ in Equation \ref{eq:temporal}. Update parameters of $g_i^{\prime}$.
    \item Obtain the final GNN ecoder $g_i^{\prime\prime}$.
\end{enumerate}

The parameters of $g$ will then be frozen to be used as a feature extractor in the downstream phishing scam detection task.

% \subsubsection{anchor nodes-based SIEGE}

\section{Experiments}
\label{sec:exp}
In this section, we evaluate the effectiveness of \sysname on Ethereum transaction data. We firstly introduce our experimental setup in Section \ref{sec:setup}, including data collection, data pre-processing, baseline models, hyperparameter selection, etc. We then present our experimental results with discussion and analysis in Section \ref{sec:result}, including both transductive and inductive settings. Furthermore, we conduct ablation study experiments in Section \ref{sec:ablation} to investigate the effectiveness of each module.

\subsection{Experimental setup}
\label{sec:setup}

\paragraph{Data collection and pre-processing.} The first block of Ethereum was generated in Jul 2015. Without loss of generality, we collected transaction data on Ethereum from January 2018 to May 2020, which is complex and contains lots of noisy information. To make experiments more clear, we list our data preparation steps as follows:

\begin{enumerate}
    \item There are two kinds of transactions in Ethereum: 1) external transactions which start from user accounts and 2) internal transactions which start from smart contract accounts. Since both 1) and 2) have a significant impact on the transaction graph, we include both of them in our dataset.
    \item We then filter out those unsuccessful transactions and zero-value transactions, which are meaningless for this task. And get a set of 75382756 transactions in total.
    \item According to block number (equivalently, timestamp), we divided the collected transaction data into five pieces and each piece was further divided into five splits. We conduct experiments with multiple pieces of graph data to reduce variability. In each piece, different graph splits can be used for incremental training.
    \item In each graph split, since no available attributes of nodes can be found in the original transaction data, and two nodes can have multiple transactions, which may lead to a multi-edge transaction graph, to simplify these issues, we use the transaction data to generate node attributes manually to get a single-edge directed graph with node attributes. Then we choose the biggest weak connected component (WCC) as the final training data. Detailed node attributes generation process can be found in Appendix A.
\end{enumerate}

%<<<<<<< HEAD
%For phishing scam nodes used for evaluation, we
%Take the first five graph splits for example, we show the data statistics in Table.
%=======
Take the first three graph splits for example, we show the data statistics in Table \ref{tab:data_statistics}. For phishing scam nodes used for evaluation, we collected labels from the Ethereum browser\footnote[1]{https://etherscan.io/} and some blacklists released by companies\footnote[2]{https://github.com/CryptoScamDB/blacklist}. In all, we obtained 6588 unique Ethereum account IDs related to phishing scams.
%>>>>>>> 73926b523ea794adf724895108b76672998897a1

\begin{table}[ht!]
  \caption{Data statistics for the first three graph splits. For overlap ratio, we refer to the proportion of overlap node set between the current and next split in the node set of the whole graph split.}
  \label{tab:data_statistics}
  \centering
  \begin{tabular}{cccccc}
    \toprule
    \textbf{Graph split}    &   \textbf{Nodes}   &   \textbf{Edges}   &   \textbf{Node features}   &   \textbf{Phishing scam nodes}   & \textbf{Overlap ratio}\\
    \midrule
   Split 1   &  2216306   &   3978100   & 17 & 114 &  0.27 \\
   Split 2   &  2201898   &   3406033   & 17 & 109 &  0.29 \\
   Split 3   &  2620644   &   4668867   & 17 & 121 &  0.26 \\
    \bottomrule
  \end{tabular}
\end{table}

\paragraph{Hyperparameter settings and baseline models.} For transductive settings, in each piece, the first three splits can be used for pre-training and incremental learning, and evaluated in the fourth split. For inductive settings, the first four splits can be used for pre-training and incremental learning. And evaluated in the fifth split (without any training). We randomly sample some nodes without phishing scam labels as normal nodes from the current graph split. In each graph split, we set the number of normal node labels to be three times that of phishing scam nodes. Then in these nodes with labels, train/val/test sets are randomly split with the ratio of 50\%/20\%/30\%.

We choose the 2-layer GraphSage\cite{hamilton2017inductive} with mean-pooling aggregator as the backbone for the node classification model. We use a logistic regression model as our binary classifier. In transductive setting, we compared SEIGE with six baselines: 1) Raw features; 2) the DeepWalk algorithm\cite{perozzi2014deepwalk}; 3) concatenation of DeepWalk node embeddings and raw features; 4) GraphSage\cite{hamilton2017inductive}, a flexible inductive GNN method; 5) DGI\cite{velivckovic2018deep}, an unsupervised inductive GNN method through maximizing the mutual information between local patch representations and high-level graph representation; 6) GCN\cite{kipf2016semi} without training, according to \cite{kipf2016semi}, GCN without training can also be considered as a powerful feature extractor. For inductive setting evaluated in an entirely new graph, the node embeddings generated by DeepWalk will become rotated with respect to the original embedding space, as pointed out in\cite{hamilton2017inductive}, so we do not compare with it in this setting. 

The Adam\cite{kingma2014adam} is used as our optimizer. We search the learning rate from $\{0.01, 0.001, 0.0001\}$. We set the dropout to 0.5. The hidden size is searched from $\{32,64,128,256\}$. For mini-batch training, the batch size is set to 512. For GraphSage, we use mean-pooling as the aggregator. For DGI, following\cite{velivckovic2018deep}, we set the two-layer GCN and three-layer GraphSage-GCN with skip connection as an encoder in transductive and inductive settings, respectively. For our model SEIGE, the value of k in the spatial pretext task is 2. And the number of graph splits used for incremental training is searched from $\{1,2,3\}$.

\subsection{Results}
\label{sec:result}

Our results for transductive and inductive settings are shown in Table \ref{tab:results_1} and Table \ref{tab:results_2}, respectively. For the transductive setting, it is obvious that SEIGE outperforms baselines with about 4\% in F-1 score, which strongly demonstrates the effectiveness of our SEIGE model. We also tried to remove the incremental learning module from SEIGE. The comparison with the original model shows that the incremental learning method is also helpful for the detection of phishing scam nodes, which actually is a general mechanism applicable to other inductive GNN methods. It can be seen that the SEIGE without incremental training could also perform better than baselines, which demonstrates the effectiveness of our SSL modules. Furthermore, we note that the DGI method slightly outperforms the GraphSage method, and that an untrained GCN also yields good results.

\begin{table}[ht!]
  \caption{Prediction results in the transductive setting in terms of \textit{accuracy}, \textit{recall}, \textit{precision}, and \textit{F1 score}. The results are averaged over 5 runs in different graph pieces. '-incremental' means that we do not use the incremental training.}
  \label{tab:results_1}
  \centering
  \begin{tabular}{lcccc}
    \toprule
    \textbf{Method}    & Acc(\%)   &  Precision(\%)    &   Recall(\%)    & F-1(\%)  \\
    \midrule
    Raw features   &  56.9 & 29.9 & 53.9 & 38.5 \\
    DeepWalk\cite{perozzi2014deepwalk}   & 60.2 & 31.8 & 51.8 & 39.4  \\
    DeepWalk+features   & 64.1 & 36.0 & 55.9 & 43.8 \\
    \midrule
    GraphSage\cite{hamilton2017inductive}   & 66.7 & 38.7 & 56.9 & 46.1  \\
    GCN-untrained\cite{kipf2016semi}   &  69.2 & 41.7 & 58.0 & 48.5 \\
    DGI\cite{velivckovic2018deep}   &  69.0 & 41.9 & 62.9 & 50.3 \\
    \midrule
    \textbf{SEIGE}   & \textbf{71.7} & \textbf{45.6} & \textbf{68.0} & \textbf{54.6} \\
    \textbf{SEIGE}-incremental   & 70.4 & 43.9 & 65.8 & 52.7 \\

    \bottomrule
  \end{tabular}
\end{table}

For inductive setting, as expected, we found that the raw feature method and the GCN-untrained method are not considerably changed compared to transductive settings. While other methods have a slight drop in performance, which may be caused by a slight change in the transaction data distribution. 

\begin{table}[ht!]
  \caption{Prediction results in the inductive setting in terms of \textit{accuracy}, \textit{recall}, \textit{precision}, and \textit{F1 score}. The results are averaged over 5 runs in different graph pieces. '-incremental' means that we do not use the incremental training.}
  \label{tab:results_2}
  \centering
  \begin{tabular}{lcccc}
    \toprule
    \textbf{Method}    & Acc(\%)   &  Precision(\%)    &   Recall(\%)    & F-1(\%)  \\
    \midrule
    Raw features   &  57.4 & 30.0 & 52.9 & 38.3 \\
    \midrule
    GraphSage\cite{hamilton2017inductive}   & 63.7 & 35.2 & 53.9 & 42.6  \\
    GCN-untrained\cite{kipf2016semi}   &  68.7 & 40.8 & 55.9 & 47.2 \\
    DGI\cite{velivckovic2018deep}   &  67.2 & 40.0 & 62.0 & 48.6 \\
    \midrule
    \textbf{SEIGE}   & \textbf{70.0} & \textbf{43.5} & \textbf{66.9} & \textbf{52.7} \\
    \textbf{SEIGE}-incremental   & 67.7 & 40.7 & 63.9 & 49.7 \\

    \bottomrule
  \end{tabular}
\end{table}

\subsection{Ablation study}
\label{sec:ablation}

We performed an ablation study for SEIGE in the transductive setting and the results have shown that both the spatial pretext task and temporal pretext task are significant for SEIGE. It can also be seen that compared to the temporal pretext task, the spatial pretext task is more important, which indicates that maybe the spatial relationship has a larger impact on this phishing scam node detection.

\begin{table}[ht!]
  \caption{Ablation study of SEIGE in the transductive setting in terms of \textit{accuracy}, \textit{recall}, \textit{precision}, and \textit{F1 score}.}
  \label{tab:ablation}
  \centering
  \begin{tabular}{lcccc}
    \toprule
    \textbf{Method}    & Acc(\%)   &  Precision(\%)    &   Recall(\%)    & F-1(\%)  \\
    \midrule
    \textbf{SEIGE}   & \textbf{71.7} & \textbf{45.6} & \textbf{68.0} & \textbf{54.6} \\
    \textbf{SEIGE}-incremental   & 70.4 & 43.9 & 65.8 & 52.7 \\
    \textbf{SEIGE}-temporal   & 65.1 & 36.8 & 55.0 & 44.1 \\
    \textbf{SEIGE}-spatial   & 60.2 & 32.5 & 55.0 & 40.9 \\
    \bottomrule
  \end{tabular}
\end{table}

\section{Conclusion}

In this paper, to solve the problem of phishing scam node detection on Ethereum transaction graph. We propose a self-supervised deep graph learning method with incremental training to address the label scarcity and the data scalability problem. In our self-supervised learning module, we design two new pretext tasks from both spatial and temporal dimensions. Our extensive experimental results on large-scale Ethereum transaction graphs show that our method strongly outperforms baseline models.

\bibliographystyle{plain}
\bibliography{neurips_2021}

\newpage
\appendix

\section{Node attributes generation}
Since the original Etheruem transaction graph does not have initial features, we explain our node attributes (initial features) generation process in detail here \ref{tab:node_attr}. We extracted features from the raw transaction data as follows,

\begin{table}[htb!]
  \caption{The node attributes (initial features) generation process. We selected 17 general features from raw Ethereum transaction data as the node attributes.}
  \label{tab:node_attr}
  \centering
  \begin{tabular}{ll}
    \toprule
    \textbf{Index}    & \textbf{Feature} (for each node)  \\
    \midrule
    \textbf{1}  & is smart contract or not.   \\
    \textbf{2}  & the in-degree.   \\
    \textbf{3}  & the out-degree.   \\
    \textbf{4}  & the number of in-transactions.   \\
    \textbf{5}  & the number of out-transactions.   \\
    \textbf{6}  & the sum of all transaction amount.   \\
    \textbf{7}  & the sum of all in-transaction amount.   \\
    \textbf{8}  & the sum of all out-transaction amount.   \\
    \textbf{9}  & the average of all in-transaction amount.   \\
    \textbf{10}  & the average of all out-transaction amount.   \\
    \textbf{11}  & the time span of all transactions.   \\
    \textbf{12}  & the time span of all in-transactions.   \\
    \textbf{13}  & the time span of all out-transactions.   \\
    \textbf{14}  & the frequenncy of all in-transactions.   \\
    \textbf{15}  & the frequenncy of all out-transactions.   \\
    \textbf{16}  & the proportion of multiple in-transactions between same node pair.   \\
    \textbf{17}  & the proportion of multiple out-transactions between same node pair.   \\
    \bottomrule
  \end{tabular}
\end{table}

\section{Hardware and further discussion}

In this section, we introduce the hardware we use in the experiments and give some further discussion about the performance bottleneck and the incremental learning module in SEIGE.

We ran our experiments in a single machine with 4 GPUs of TITAN Xp  (12Gb of RAM), 12 CPUs of Intel Core i7-6850K (@ 3.60GHz). OS version: Ubuntu 16.04.4 LTS, and CUDA version: 11.0.

One of the main performance bottlenecks for the phishing scam detection task is the label scarcity, which is why we are using unsupervised learning. If we can obtain more labels on Ethereum for training then we may be able to achieve better results. 

In addition, we would like to further discuss the incremental learning module in SEIGE. First, this incremental learning module works well to alleviate the problem of data scalability. In fact, in our experiments, for deep graph model baselines: GraphSage, DeepWalk, and DGI, if we combine two graph splits into one and use it for training, it leads to out of memory problem. Moreover, the incremental learning module is a general mechanism that can be used in other deep graph models as well.

\section{Visualization}

\begin{figure}[ht!]
    \centering
    \includegraphics[width=1.0\linewidth]{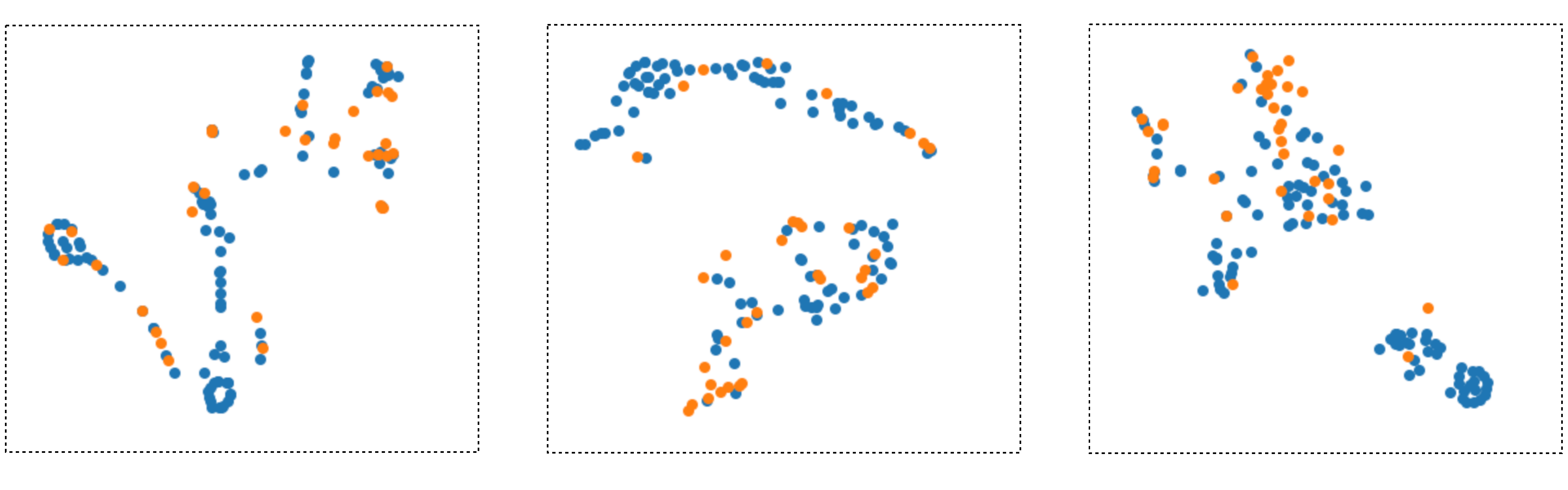}
    \caption{Node features visualization of raw features (\textbf{left}), GCN-untrained (\textbf{middle}), and SEIGE (\textbf{right}) by t-SNE\cite{van2008visualizing}. Orange points stand for phishing scam nodes and blue points are normal nodes. To make the figure clear, we limit the number of points to about 500 and the ratio of normal nodes to phishing nodes is about 4:1.}
    \label{fig:vis}
\end{figure}

We show the node features visualization of three methods in Figure \ref{fig:vis}: i) raw features, ii) GCN-untrained, iii) SEIGE. We can find that points are better separated in GCN-untrained and SEIGE than raw features. Besides, we can also notice that the points in SEIGE are better separated and more concentrated compared to GCN-untrained. It further demonstrates the effectiveness of our SEIGE model.

\end{document}